# Spatial-Temporal Digital Image Correlation: A Unified Framework

Yuxi Chi, Bing Pan*

Institute of Solid Mechanics, School of Aeronautic Science and Engineering, Beihang University, Beijing 100191, China

Email: panb@buaa.edu.cn

**Abstract:** A comprehensive and systematic framework for easily extending and implementing the subset-based spatial-temporal digital image correlation (DIC) algorithm is presented. The framework decouples the three main factors (i.e. shape function, correlation criterion, and optimization algorithm) involved in algorithm implementation of DIC and represents different algorithms in a uniform form. One can freely choose and combine the three factors to meet his own need, or freely add more parameters to extract analytic results. Subpixel translation and a simulated image series with different velocity characters are analyzed using different algorithms based on the proposed framework, confirming the merit of noise suppression and velocity compatibility. An application of mitigating air disturbance due to heat haze using spatial-temporal DIC is given to demonstrate the applicability of the framework.

## 1  Introduction

Digital image correlation (DIC) is a non-contact image-based optical method for full-field shape, displacement and strain measurements. After the continual improvements made during the last three decades, DIC techniques have been widely accepted as the most popular, practical and powerful optical metrology in Experimental Mechanics community [1-3]. The basic idea of the subset-based local DIC is simple and straightforward, namely the registering (matching or tracking) the corresponding points in the reference and deformed images yields the surface displacement and strain fields of a test object. To this end, a square subset with proper size should first be specified at each measurement point. Then, an objective function (also known as correlation criterion) combined with a proper shape function should be defined to quantify the similarity (or dissimilarity) degree between the subsets. Next, the correlation criterion is optimized by using a nonlinear optimization algorithm to extract the deformation parameters that depict the position and shape of target subsets. To achieve the purpose of high-accuracy and high-efficiency DIC measurements in various conditions, numerous efforts have been dedicated to the three main factors in DIC algorithms such as shape functions [4,5], correlation criteria [6-8], and optimization algorithms [9-12]. The three main factors are then combined into various DIC methods.

The existing DIC method performs well in most cases. However, with the increasing demand, more and more challenging scenarios are to be measured, including dynamic testing [13-15], object tracking and high-temperature situation [16-20], etc. In these situations, the noise due to high sampling frequency or thermal radiation is much higher than that in normal condition, thus leading to increase the measurement error or even a failure of DIC analysis. For example, in dynamic testing, the velocity and strain rate are usually much higher, which may make DIC fail in initial guess stage. In high-temperature applications, the virtual deformation due to air disturbance fluctuates with time and couples with true deformation, which makes the results unreliable. It can be found that these challenges are associated with time, thus, the temporal constraint is the key to these problems. In DIC, the shape function enforces the spatial continuity within a subset, which follows the basic assumption of continuum mechanics. However, as time flows smoothly, the continuity also exists in the temporal domain. We can follow the same idea that use the temporal shape function (TSF) to describe the deformation along time. Also, different temporal components should be considered according to the form of the deformation rate. The temporal constraint will help to reduce the adverse effects of disturbances (noise, air disturbance, etc.) that varies randomly over time. Further, TSF considers the images before and after the current one, which means more information is included in the correlation analysis. Consequently, results with higher accuracy are to be expected.

Many researchers have discussed the temporal components. Broggiato et.al. [21] firstly considered the temporal components in strain rate measurement, which used the second-order TSF that fit the pixel displacements in five consecutive images into a parabola. The temporal constraint can deal with general kinds of motion. Also, different velocity characteristics [22] (constant, parabolic and discontinuous) were discussed. The idea that integrating temporal components was quickly adopted by the community of global DIC [23] and integrated DIC [24]. Efforts were also made to subset-based DIC. For instance, Wang et.al. [25] discussed the first order TSF with the assumption of the constant velocity. Different from Broggiato's work that updates the parameters in all five images, Wang's method used the average of consecutive images as the deformed image and calculate the displacement of the current subset in the middle image. Combining the fast subset-based DIC with temporal components, the method shows its merit on calculation speed over 4-node-quadrangle-element DIC (T-Q4-DIC) while keeping the noise suppression ability. In Wang's work, 1-order SF and TSF, Newton-Raphson (NR) and sum of squared difference (SSD) were used to build the spatial-temporal DIC algorithm. However, the formulas were deduced in a coupled form. When more parameters (e.g. higher order, strain rates) to be considered, or different correlation criteria or new optimization algorithms to be used, a decoupled form and a uniform framework that can be readily used to establish corresponding spatial-temporal DIC (ST-DIC) would be a great help.

In this work, we proposed a unified framework for spatial-temporal DIC based on the least square principle. In this framework, temporal and spatial parameters are treated as equal status, which enables the use of efficient IC-GN algorithm. The influences of three factors (i.e., shape function parameters, correlation criterion and optimization algorithm) mentioned above are decoupled and detailed in this work and it can serve as a guide for researchers to explore new algorithms. We will start with the basic least squares principles and show how to build a simple DIC algorithm from scratch. The three factors are then analyzed and added to the framework individually. The unified shape functions are analyzed and discussed through several examples. The common parts and differences between temporal and spatial dimension are also analyzed. With the help of the given tutorial, new methods can be easily implemented based on this framework. The merits of noise suppressing, better initial guessing and air disturbance reducing of ST-DIC are demonstrated through several experiments. The rest parts are organized as follows. Section 2 details the proposed framework and gives a tutorial of building a customized DIC algorithm.

Then, two simulated experiments and a high-temperature one are shown in section 3. Finally, we give the conclusion in section 4.

## 2 Principles

### 2.1 DIC: A special least squares (LSQ) problem

Mathematically, least squares is used to find the best approximation that minimizes the sum of squared differences (SSD) between the true value and the parametric modeled value. Without loss of generality, let $\mathbf{p} \in \mathbb{R}^n$ be an n-dimensional parameter vector, and $\mathbf{r}(\mathbf{p}) = [r_1(\mathbf{p}), r_2(\mathbf{p}), \ldots, r_m(\mathbf{p})]^T$ the m-dimensional residual vector. The least squares problem is expressed as:

$$arg \min_{\mathbf{p} \in \mathbb{R}^n} f(\mathbf{p}) = \frac{1}{2} \| \mathbf{r}(\mathbf{p}) \|^2 = \frac{1}{2} \mathbf{r}(\mathbf{p})^T \mathbf{r}(\mathbf{p}) \tag{1}$$

In the expression, we can see that the LSQ problem is only related to the form of the residual vector $\mathbf{r}(\mathbf{p})$. The detailed solution of LSQ problem can be found in Appendix A.

As mentioned before, DIC needs to establish a similarity measure that combines a correlation criterion and a shape function to model the deformation between the reference and the deformation subsets. It aims to optimize the parameters in shape function to minimize the difference between the modeled deformed image and the true counterparts. Therefore, DIC can be regarded as a special LSQ. Here, we rewrite the DIC problem in the form of LSQ and build a straightforward DIC algorithm from scratch.

Consider a subset with the center located at $[x_C, y_C]^T$. The first-order shape function with parameters $\mathbf{p}$ map the pixel at $\mathbf{X} = [x, y]^T$ to the modeled location as:

$$\tilde{\mathbf{X}}(\mathbf{p}) = \begin{bmatrix} \tilde{x} \\ \tilde{y} \end{bmatrix} = \mathbf{X} + \begin{bmatrix} u & u_x & u_y \\ v & v_x & v_y \end{bmatrix} \begin{bmatrix} 1 \\ \Delta x \\ \Delta y \end{bmatrix} \tag{2}$$

where $\Delta x = x - x_C$, $\Delta y = y - y_C$, $\mathbf{p} = [u, u_x, u_y, v, v_x, v_y]^T$.

Here we denote the reference subset as **F** and the deformation subset as **G**, and use the difference of grayscale between the grayscale of reference pixel and that in the deformed image at the modeled location as the residual. The residual component at **X** can be expressed as:

$$\mathbf{r}(\mathbf{p})_\mathbf{X} = \mathbf{F}(\mathbf{X}) - \mathbf{G}(\tilde{\mathbf{X}}(\mathbf{p})) \qquad (3)$$

As in (A.8), the key to solve LSQ problem is the Jacobi matrix. Here we write the Jacobi vector at **X** as:

$$\begin{aligned}\mathbf{J}(\mathbf{p})_\mathbf{X} &= \frac{\partial \mathbf{r}(\mathbf{p})_\mathbf{X}}{\partial \mathbf{p}} = \frac{\partial}{\partial \mathbf{p}}\left(\mathbf{F}(\mathbf{X}) - \mathbf{G}(\tilde{\mathbf{X}}(\mathbf{p}))\right) \\ &= -\frac{\partial \mathbf{G}}{\partial \tilde{\mathbf{X}}} \frac{\partial \tilde{\mathbf{X}}}{\partial \mathbf{p}} \\ &= -\begin{bmatrix} G_{\tilde{x}} \\ G_{\tilde{y}} \end{bmatrix}^T \begin{bmatrix} 1 & \Delta\tilde{x} & \Delta\tilde{y} & 0 & 0 & 0 \\ 0 & 0 & 0 & 1 & \Delta\tilde{x} & \Delta\tilde{y} \end{bmatrix}\end{aligned} \qquad (4)$$

where $G_x$ and $G_y$ are derivatives of the deformation subset **G** w.r.t. $x$ and $y$, respectively.

For every pixel located at **X** in the subset, we can write the corresponding Jacobi vector $\mathbf{J}(\mathbf{p})_\mathbf{X}$. This is a row vector of 6 elements, and we can vertically stack the Jacobi vectors to the Jacobi matrix $\mathbf{J}(\mathbf{p})$. For a subset with $n \times n$ pixels, the shape of the Jacobi matrix will be $n^2$-by-6.

Using (A.9), the same iterative form can be established as:

$$\mathbf{p}^{(k+1)} = \mathbf{p}^{(k)} - \mathbf{J}\left(\mathbf{p}^{(k)}\right)^\dagger \mathbf{r}\left(\mathbf{p}^{(k)}\right) \qquad (5)$$

In this example, a simple first-order shape function and a simple sum of squared difference (SSD) criterion are used to describe the DIC problem, and the Gauss-Newton method is used to solve the parameters. As seen above, the DIC is a special case of nonlinear LSQ. The model function in DIC is the shape function. The residual vector in DIC is the grayscale difference between the reference subset and the modeled deformation subset while in general

LSQ, the residual vector is the true value with the modeled one. What makes DIC different is the image that serves as a function mapping the location to grayscale. It makes the comparison happen on the grayscale rather than the location. As we can see in the expression of Jacobi vector (4), the derivatives of the parameters are propagated through the shape function to the location, then go through image gradient to the grayscale.

## 2.2 Common part of spatial DIC and ST-DIC

### 2.2.1 Optimization algorithm

In DIC, considered its special characteristics that the two images have equal status in the matching problem, methods can be divided into forward and inverse methods according to the role of reference. Also, methods can be divided into additive and compositional methods according to the way of parameters updating. Further, to determine the increment of parameters, there are countless mathematical methods such as Newton-Raphson, Gauss-Newton, Levenberg-Marquardt, trust region, dogleg, and many other methods. These traits can be freely combined into an optimization algorithm. The mostly used forward-additive Newton-Raphson (FA-NR) method [9] and inverse-compositional Gauss-Newton (IC-GN) method [12] can be easily built on the proposed framework. The simplest and most intuitive FA method has been detailed in 2.1 and the compositional methods are described below.

In additive methods, parameters are updated through a simple addition operation. However, compositional methods are rather different from the additive ones. In compositional methods, the first order shape function is written in the homogeneous form:

$$\begin{bmatrix} 1 \\ \tilde{x} - x_C \\ \tilde{y} - y_C \end{bmatrix} = \begin{bmatrix} 1 & 0 & 0 \\ u & 1+u_x & u_y \\ v & v_x & 1+v_y \end{bmatrix} \begin{bmatrix} 1 \\ x - x_C \\ y - y_C \end{bmatrix} \qquad (6)$$

then, in the local coordinate as:

$$\tilde{\mathbf{X}}_H = \begin{bmatrix} 1 \\ \tilde{x} \\ \tilde{y} \end{bmatrix} = \begin{bmatrix} 1 & 0 & 0 \\ u & 1+u_x & u_y \\ v & v_x & 1+v_y \end{bmatrix} \begin{bmatrix} 1 \\ x \\ y \end{bmatrix} \qquad (7)$$
$$= \mathbf{W}(\mathbf{p})\mathbf{X}_H$$

where $\mathbf{X}_H$ is the local coordinates in the homogeneous form.

Since the first order shape function is an affine transformation and all the invertible affine transformations form an affine group, the increment $\Delta \mathbf{p}$ can be composited with the current parameter $\mathbf{p}$ to get an updated warp matrix. As described in[10,11], the increment $\Delta \mathbf{p}$ warps the location $\mathbf{X}_H$ in the warped image to $\mathbf{W}(\Delta \mathbf{p})\mathbf{X}_H$, thus the composited warp matrix transform the location $\mathbf{X}_H$ in the original image to $\mathbf{W}(\mathbf{p})\mathbf{W}(\Delta \mathbf{p})\mathbf{X}_H$. That is, the parameters are updated as:

$$\mathbf{W}(\mathbf{p}^{(k+1)}) = \mathbf{W}(\mathbf{p}^{(k+1)})\mathbf{W}(\Delta \mathbf{p}) \qquad (8)$$

In light of (8), the residual component at $\mathbf{X}_H$ can be expressed as:

$$\mathbf{r}(\mathbf{p})_{\mathbf{X}_H} = \mathbf{F}(\mathbf{X}_H) - \mathbf{G}(\mathbf{W}(\mathbf{p})\mathbf{W}(\Delta \mathbf{p})\mathbf{X}_H) \qquad (9)$$

and the Jacobi vector at $\mathbf{X}$ is expressed as:

$$\mathbf{J}(\mathbf{p})_{\mathbf{X}_H} = \frac{\partial}{\partial \Delta \mathbf{p}}\left(\mathbf{F}(\mathbf{X}_H) - \mathbf{G}(\mathbf{W}(\mathbf{p})\mathbf{W}(\Delta \mathbf{p})\mathbf{X}_H)\right)$$
$$= -\frac{\partial \mathbf{G}}{\partial \mathbf{W}(\mathbf{p})\mathbf{X}_H} \frac{\partial \mathbf{W}(\mathbf{p})\mathbf{X}_H}{\partial \mathbf{p}} \qquad (10)$$

where $\mathbf{W}(\mathbf{p})\mathbf{X}_H$ is the warped location, namely $\tilde{\mathbf{X}}_H$ in (4). That is, in forward compositional method, the increment $\Delta \mathbf{p}$ is calculated in the same way as (5), but parameters are updated using the formula (8).

In forward methods, the increment is added to the warped subset. Hence, the Jacobi vector uses the gradient of the warped image, which requires to calculate the gradient of the warped image at the warped local coordinates in each iteration, as shown in (4). Because the compositional method is equivalent to the additive methods in first order approximation, as shown in (4) and (10), both additive and compositional methods have the same overheads.

In inverse compositional methods, the increment $\Delta \mathbf{p}$ is added to the reference. We write the residual component at $\mathbf{X_H}$ in the same way as:

$$\mathbf{r}(\mathbf{p})_{\mathbf{X_H}} = \mathbf{F}(\mathbf{W}(\Delta \mathbf{p})\mathbf{X_H}) - \mathbf{G}(\mathbf{W}(\mathbf{p})\mathbf{X_H}) \tag{11}$$

then the chain rule gives the Jacobi vector at $\mathbf{X_H}$:

$$\begin{aligned} \mathbf{J}(\mathbf{p})_{\mathbf{X_H}} &= \frac{\partial}{\partial \Delta \mathbf{p}} \left( \mathbf{F}(\mathbf{W}(\Delta \mathbf{p})\mathbf{X_H}) - \mathbf{G}(\mathbf{W}(\mathbf{p})\mathbf{X_H}) \right) \\ &= \frac{\partial \mathbf{F}}{\partial \mathbf{X_H}} \frac{\partial \mathbf{W}(\Delta \mathbf{p}) \mathbf{X_H}}{\partial \Delta \mathbf{p}} \\ &= \begin{bmatrix} F_x \\ F_y \end{bmatrix}^T \begin{bmatrix} 1 & x & y & 0 & 0 & 0 \\ 0 & 0 & 0 & 1 & x & y \end{bmatrix} \\ &= \begin{bmatrix} F_x \\ F_y \end{bmatrix}^T \begin{bmatrix} \mathbf{X_H}^T & \mathbf{0} \\ \mathbf{0} & \mathbf{X_H}^T \end{bmatrix} \end{aligned} \tag{12}$$

where the increment $\Delta \mathbf{p}$ is calculated using (A.8) at local coordinates, not the warped one.

Then the increment is to be composited with the current parameters $\mathbf{p}$. Let $\mathbf{X_H}' = \mathbf{W}(\Delta \mathbf{p})\mathbf{X_H}$ be the warped location, $\mathbf{X_H}$ will be $\mathbf{X_H} = \mathbf{W}(\Delta \mathbf{p})^{-1} \mathbf{X_H}'$ and the residual component can therefore be expressed as:

$$\begin{aligned} \mathbf{r}(\mathbf{p})_{\mathbf{X_H}} &= \mathbf{F}(\mathbf{W}(\Delta \mathbf{p})\mathbf{X_H}) - \mathbf{G}(\mathbf{W}(\mathbf{p})\mathbf{X_H}) \\ &= \mathbf{F}(\mathbf{X_H}') - \mathbf{G}(\mathbf{W}(\mathbf{p})\mathbf{W}(\Delta \mathbf{p})^{-1}\mathbf{X_H}') \end{aligned} \tag{13}$$

Compare (13) with (11), it is clear that the parameters are updated by the following formula:

$$\mathbf{W}(\mathbf{p}^{(k+1)}) = \mathbf{W}(\mathbf{p}^{(k)})\mathbf{W}(\Delta \mathbf{p})^{-1} \tag{14}$$

Note that when the inverse compositional method is used, we can only update the rows w.r.t. *x* and *y*. The two rows contain all the parameters, and the two rows in the updated warp matrix depend only on the two rows in the current warp matrix. This trick can further improve the calculation efficiency.

In inverse compositional method, the increment is added to the reference subset. Consequently, the Jacobi is independent of the warped subset as shown in (12) and only the gradient of the reference image is used. The Jacobi vector is only the function of the reference gradient and the pixel coordinates and has no connection with the parameters **p**. This feature allows us to calculate the Jacobi and its pseudo-inverse in advance without repeated calculations in the iterations as in forward methods, thus greatly improving the calculation speed and making IC-GN to be the defaco standard algorithm in DIC.

### 2.2.2 Correlation criterion

In LSQ problem, the correlation criterion or object function is simply selected as the L2-Norm of the residual vector based on the least square principle. In the development history of DIC, various criteria have been proposed. In early days, simple and intuitive criteria are often selected, including cross-correlation (CC), sum of squared difference (SSD), namely the L2-Norm, and parametric sum of squared difference (PSSD). Considering the robustness, researchers proposed the normalized criteria, including zero-mean cross-correlation (ZNCC), zero-mean normalized sum of squared difference (ZNSSD), and parametric sum of square difference (PSSDab). As has been proved by Pan et.al[26], these normalized criteria are equivalent. As the simplified version, those criteria without normalization are also equivalent. Here, we just take SSD and ZNSSD for comparison and show how to use ZNSSD in this framework.

ZNSSD consider the offset and scale changes of the grayscale by subtracting the mean value then rescale to norm one. Mathematically, ZNSSD between two images **F** and **G** is:

$$\text{ZNSSD}(\mathbf{F},\mathbf{G}) = \left\| \frac{\mathbf{F}-\overline{\mathbf{F}}}{\left\|\mathbf{F}-\overline{\mathbf{F}}\right\|} - \frac{\mathbf{G}-\overline{\mathbf{G}}}{\left\|\mathbf{G}-\overline{\mathbf{G}}\right\|} \right\|^2$$
$$= \frac{1}{n^4} \left\| \frac{\mathbf{F}-\overline{\mathbf{F}}}{\Delta\mathbf{F}} - \frac{\mathbf{G}-\overline{\mathbf{G}}}{\Delta\mathbf{G}} \right\|^2 \tag{15}$$

where $\Delta\mathbf{F}$ is the standard derivation of the grayscales in the subset, $n$ is the subset size. Since the constant scaling factor $1/n^4$ has no effect on optimization. The residual component at $\mathbf{X}$ can therefore be expressed as:

$$\mathbf{r}(\mathbf{p})_\mathbf{X} = \frac{\mathbf{F}(\mathbf{X}) - \overline{\mathbf{F}}}{\Delta \mathbf{F}} - \frac{\mathbf{G}(\tilde{\mathbf{X}}(\mathbf{p})) - \overline{\mathbf{G}}}{\Delta \mathbf{G}} \tag{16}$$

With the residual expression, we can easily port algorithms using SSD to ZNSSD. Here we take the classic FA-NR (4) and IC-GN (12) as examples. For FA-NR method, calculating the Jacobi vector at $\mathbf{X}$ using the chain rule, we can obtain a similar expression:

$$\mathbf{J}(\mathbf{p})_\mathbf{X} = \frac{\partial \mathbf{r}(\mathbf{p})_\mathbf{X}}{\partial \mathbf{p}} = -\frac{1}{\Delta \mathbf{G}} \frac{\partial \mathbf{G}}{\partial \tilde{\mathbf{X}}} \frac{\partial \tilde{\mathbf{X}}}{\partial \mathbf{p}} \tag{17}$$

Compare (17) with (4), we can find that the Jacobi vector is just proportional to that in SSD criterion, also the pseudo-inverse of Jacobi matrix and the updating parameter vector.

$$\begin{aligned} \mathbf{J}_{ZNSSD}(\mathbf{p})_\mathbf{X} &= \frac{1}{\Delta \mathbf{G}} \mathbf{J}_{SSD}(\mathbf{p})_\mathbf{X} \\ \mathbf{J}_{ZNSSD}(\mathbf{p})^\dagger &= \Delta \mathbf{G} \mathbf{J}_{SSD}(\mathbf{p})^\dagger \\ \Delta \mathbf{p} &= -\Delta \mathbf{G} \mathbf{J}_{SSD}(\mathbf{p})^\dagger \mathbf{r}(\mathbf{p}) \end{aligned} \tag{18}$$

Similarly, in IC-GN method, the Jacobi vector at $\mathbf{X}_\mathbf{H}$ is expressed as:

$$\mathbf{J}(\mathbf{p})_{\mathbf{X}_\mathbf{H}} = \frac{1}{\Delta \mathbf{F}} \frac{\partial \mathbf{F}}{\partial \mathbf{X}_\mathbf{H}} \frac{\partial \mathbf{W}(\Delta \mathbf{p}) \mathbf{X}_\mathbf{H}}{\partial \Delta \mathbf{p}} \tag{19}$$

where the only difference is the proportional factor changed from $\Delta \mathbf{G}$ to $\Delta \mathbf{F}$. The similar connections between SSD and ZNSSD in IC-GN method is then shown below:

$$\Delta \mathbf{p} = -\Delta \mathbf{F} \mathbf{J}_{SSD}(\mathbf{p})^\dagger \mathbf{r}(\mathbf{p}) \tag{20}$$

We have seen how to replace the criterion using ZNSSD while keeping the rest part (shape function and optimization algorithm) unchanged. When the correlation criterion is changed from SSD to ZNSSD, the Jacobi matrix can be reused. We only need to calculate the standard deviation and the residual vector in the subset. Note that the residual vector is different from that in SSD criterion while the Hessian matrix and Jacobi matrix are the same.

## 2.3 Unified spatial and temporal shape function

In DIC that does not consider the temporal components, only spatial parameters are optimized. Different shape functions (SF) are used to describe the deformations. The zero-order SF considering the translation has only two parameters, i.e. $u$ and $v$. The first-order SF consider the translation and affine transformation, which add the first-order spatial derivatives $(u_x, u_y, v_x, v_y)$. The second-order SF consider higher order of deformation, and second-order spatial derivatives $(u_{xx}, u_{xy}, u_{yy}, v_{xx}, v_{xy}, v_{yy})$ are added. Also, any other parameters that can describe deformation can be added into the SF, that is the main idea of the integrated DIC (i-DIC)[27]. These parameters are introduced using the Taylor's series. For example, the cross item $u_{xy}$ and $v_{xy}$ can be introduced using Taylor's series as:

$$\tilde{\mathbf{X}}(\mathbf{p}) = \begin{bmatrix} \tilde{x} \\ \tilde{y} \end{bmatrix} = \mathbf{X} + \begin{bmatrix} u & u_x & u_y & u_{xy} \\ v & v_x & v_y & v_{xy} \end{bmatrix} \begin{bmatrix} 1 \\ \Delta x \\ \Delta y \\ \Delta x \Delta y \end{bmatrix} \tag{21}$$

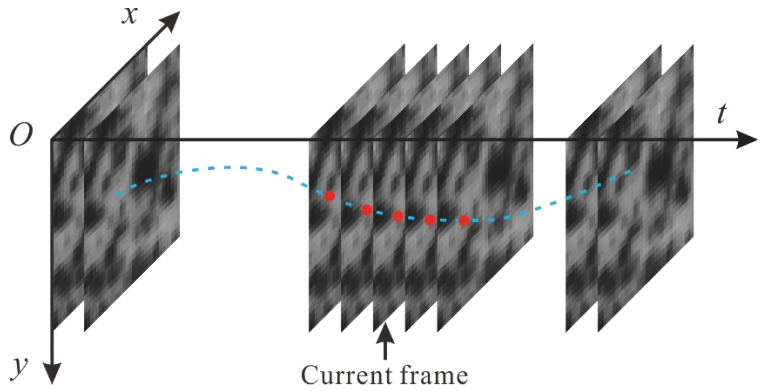

**Fig. 1** Temporal continuity considered in ST-DIC

When the temporal dimension is considered in DIC, a series of images should be analyzed. As shown in **Fig. 1**, the central frame is chosen as the calculated one, and its neighbor frames as the temporal information provider.

When temporal components are considered, there also comes the time shape function (TSF). Similarly, zero-order TSF (no additional parameters) implies the constant location between the image series while first-order TSF (add $u_t$, $v_t$) describes the constant velocity. **Fig. 2** gives the intuitive explanation of the first order temporal parameters. Also, second-order TFS (add $u_{tt}$, $v_{tt}$) can describe more complex movements. If we use the first order spatial and temporal shape function, the modeled displacement can be expressed using Taylor's series as:

$$\tilde{\mathbf{X}}(\mathbf{p}) = \begin{bmatrix} \tilde{x} \\ \tilde{y} \end{bmatrix} = \mathbf{X} + \begin{bmatrix} u & u_x & u_y & u_t \\ v & v_x & v_y & v_t \end{bmatrix} \begin{bmatrix} 1 \\ \Delta x \\ \Delta y \\ \Delta t \end{bmatrix} \tag{22}$$

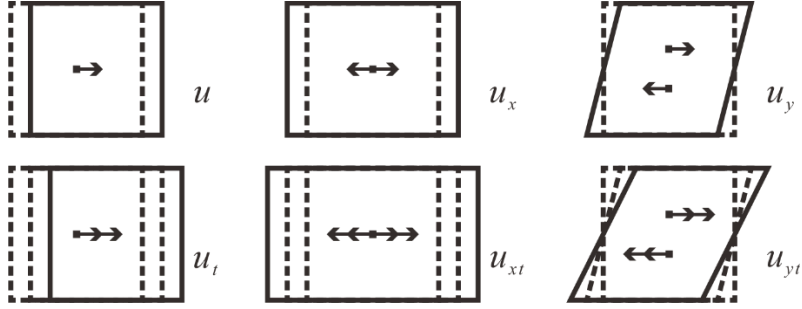

**Fig. 2** Schematic diagram of first-order temporal parameters

As has been shown, Taylor's series can easily give shape function of any order. The core idea of this framework is to treat the time and the space dimension as the same status and denotes one point using its spatiotemporal coordinate $[x, y, t]$. If image series is captured with an equal time step, the temporal coordinate can be simplified to integers. Because of the orthogonality in each dimension, same parameters w.r.t each dimension are used to describe the deformation in the shape function. Hence, the shape function of each dimension can be expressed in the same form. For each dimension, the deformation can be expressed in a uniformed way as:

$$\tilde{\mathbf{X}}_i(\mathbf{P}) = \mathbf{X}_i + \mathbf{p}_i^\mathrm{T} \mathbf{X}_\mathbf{H} \tag{23}$$

where $\mathbf{X}_i$ represent the coordinate of each dimension (x and y), $\mathbf{p}_i$ represents the parameters associated with the dimension, and $\mathbf{X}_H$ is the spatiotemporal coordinate of the current calculation point.

However, there is a small difference between the temporal and spatial dimension. When the temporal dimension is added, images are captured individually. The temporal coordinates of every image are known and keep constant. There are no parameters to describe the change of time. Hence, the temporal dimension has no contribution to the Jacobi of TSF. Apart from this, the spatial and temporal dimension can be unified in this framework.

In the deduction of the aforementioned methods, only first-order SF is shown for simplicity. In fact, these formulas do not depend on the specific form of shape functions. As shown in (17) and (19), the Jacobi consists of three terms: the factor of correlation criteria, image gradient and Jacobi of shape function. The form of shape function only affects the last term. This term propagates the derivatives from parameters to the location. They share the same term of $\partial \tilde{\mathbf{X}}/\partial \mathbf{p}$. For each dimension, the term can be simply deduced from (23) as:

$$\frac{\partial \tilde{\mathbf{X}}_i}{\partial \mathbf{p}_i} = \mathbf{X}_\mathbf{H}^\mathrm{T} \tag{24}$$

Note that the expression is independent of the specific form of shape function. That is, the shape function can be arbitrarily chosen according to the deformation form while the Jacobi shares the same form as (24). In DIC, there are two spatial dimensions X and Y. Then the Jacobi of all parameters can be just assembled as follows:

$$\frac{\partial \tilde{\mathbf{X}}}{\partial \mathbf{p}} = \begin{bmatrix} \mathbf{X}_\mathbf{H}^\mathrm{T} & \mathbf{0} \\ \mathbf{0} & \mathbf{X}_\mathbf{H}^\mathrm{T} \end{bmatrix} \tag{25}$$

Following the idea, the algorithm can be easily extended to the 3D form of DIC, known as Digital Volume Correlation (DVC). For the first order shape function in DVC, $\mathbf{X_H} = [1, \Delta x, \Delta y, \Delta z]^\mathrm{T}$, and $\mathbf{p_u} = [u, u_x, u_y, u_z]^\mathrm{T}$, $\mathbf{p_v} = [v, v_x, v_y, v_z]^\mathrm{T}$, $\mathbf{p_w} = [w, w_x, w_y, w_z]^\mathrm{T}$. Without deduction, the Jacobi of shape function at each voxel can be directly written in the same form as:

$$\frac{\partial \tilde{\mathbf{X}}}{\partial \mathbf{p}} = \begin{bmatrix} \mathbf{X}_\mathbf{H}^\mathrm{T} & 0 & 0 \\ 0 & \mathbf{X}_\mathbf{H}^\mathrm{T} & 0 \\ 0 & 0 & \mathbf{X}_\mathbf{H}^\mathrm{T} \end{bmatrix} \tag{26}$$

The uniformed form of spatial and temporal shape function has been shown. The basic idea of ST-DIC is also similar to the traditional one. The aim of ST-DIC is to minimize the difference between the reference and the warped image series. The matching is performed with each image in the series, which adds another dimension to the residual vector. The residual vector is stacked with the $n^2 m$ residual, where $m$ is the number of the image series. The residual vector at $\mathbf{X}$ and $t$ is expressed as:

$$\mathbf{r}(\mathbf{p})_{\mathbf{X},t} = \mathbf{F}(\mathbf{X}) - \mathbf{G}_t(\tilde{\mathbf{X}}(\mathbf{p})) \tag{27}$$

where $\mathbf{G}_t$ is the deformation image at time $t$.

The Jacobi vector at $\mathbf{X}$ and $t$ in the additive method are identically expressed as:

$$\mathbf{J}(\mathbf{p})_{\mathbf{X},t} = -\begin{bmatrix} G_{t\tilde{x}} \\ G_{t\tilde{y}} \end{bmatrix}^\mathrm{T} \begin{bmatrix} \tilde{\mathbf{X}}_\mathbf{H}^\mathrm{T} & 0 \\ 0 & \tilde{\mathbf{X}}_\mathbf{H}^\mathrm{T} \end{bmatrix} \tag{28}$$

and in the inverse compositional method as:

$$\mathbf{J}(\mathbf{p})_{\mathbf{X},t} = \begin{bmatrix} F_x \\ F_y \end{bmatrix}^\mathrm{T} \begin{bmatrix} \mathbf{X}_\mathbf{H}^\mathrm{T} & 0 \\ 0 & \mathbf{X}_\mathbf{H}^\mathrm{T} \end{bmatrix} \tag{29}$$

## 2.4 A tutorial on building a customized ST-DIC algorithm

The three factors of DIC have been discussed individually. Here, a tutorial on building a customized ST-DIC algorithm using the proposed framework is given as follows. In this ST-DIC algorithm, the velocity and strain rate components are considered. Also, IC-GN algorithm and ZNSSD are used to get a fast and robust ST-DIC algorithm.

**Step 1:** Choose a shape function.

The shape function including the velocity and strain rate components can be written with the help of Taylor's series:

$$\tilde{\mathbf{X}}(\mathbf{p}) = \begin{bmatrix} \tilde{x} \\ \tilde{y} \end{bmatrix} = \mathbf{X} + \begin{bmatrix} u & u_x & u_y & u_t & u_{xt} & u_{yt} \\ v & v_x & v_y & v_t & v_{xt} & v_{yt} \end{bmatrix} \begin{bmatrix} 1 \\ \Delta x \\ \Delta y \\ \Delta t \\ \Delta x \Delta t \\ \Delta y \Delta t \end{bmatrix} \quad (30)$$

Because IC-GN algorithm is used, the shape function should be written in the local and homogeneous form, and the cross terms are expanded as:

$$\begin{aligned} \tilde{x}t &= xt + ut + u_x xt + u_y yt + u_{xt} xt^2 + u_{yt} yt^2 \\ \tilde{y}t &= yt + vt + v_x xt + v_y yt + v_{xt} xt^2 + v_{yt} yt^2 \end{aligned} \quad (31)$$

Omit the high order terms and rewrite the shape function in the homogeneous form as:

$$\begin{bmatrix} 1 \\ \tilde{x} \\ \tilde{y} \\ t \\ \tilde{x}t \\ \tilde{y}t \end{bmatrix} = \begin{bmatrix} 1 & 0 & 0 & 0 & 0 & 0 \\ u & 1+u_x & u_y & u_t & u_{xt} & u_{yt} \\ v & v_x & 1+v_y & v_t & v_{xt} & v_{yt} \\ 0 & 0 & 0 & 1 & 0 & 0 \\ 0 & u & 0 & 0 & 1+u_x & u_y \\ 0 & v & 0 & 0 & v_x & 1+v_y \end{bmatrix} \begin{bmatrix} 1 \\ x \\ y \\ t \\ xt \\ yt \end{bmatrix} \quad (32)$$

$$\tilde{\mathbf{X}}_H(\mathbf{p}) = \mathbf{W}(\mathbf{p}) \mathbf{X}_H$$

**Step 2:** Pre-compute the Jacobi and Hessian matrices.

Because IC-GN algorithm is used, the image gradient and the Jacobi can be calculated in advance following (29). Also, the term $\Delta F$ of each subset can be also computed in advance.

**Step 3:** Calculate the updating parameter vector.

The updating parameter $\Delta \mathbf{p}$ should be calculated using (20), where the residual vector should be calculated using (16) since ZNSSD is used. In the formula, m images around the current calculation image are taken from the deformation image series. And the warped deformation subsets are interpolated using the m images and the warped locations with the help of shape function. Then, the warped subsets are used to calculate the residual vector.

**Step 4:** Update the parameters and perform iteration.

The parameters are updated using (14) with the calculated updating vector. Then back to step-3 and iterate until convergence.

As we can see, the tutorial is quite simple. One just needs to choose the three factors of DIC and follow the tutorial, a customized ST-DIC algorithm can be built with minimal efforts. All the formulas are in the same form.

## 3 Experiments

### 3.1 Simulation Experiment

#### 3.1.1 Sub-pixel translation analysis

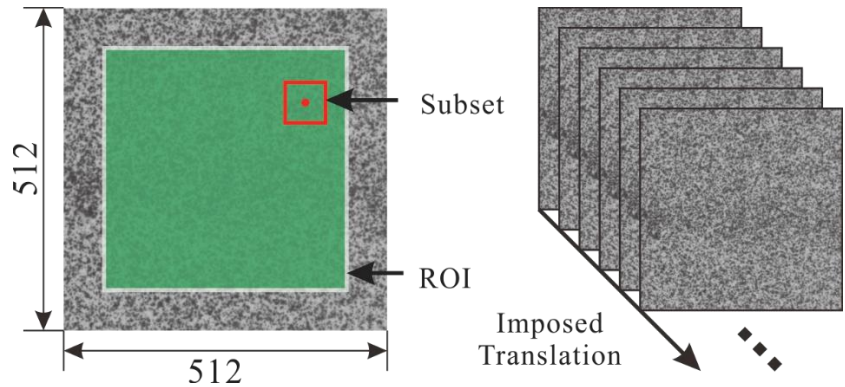

**Fig. 3** The speckle pattern used in simulated translation experiments.

To study the influence of the addition of temporal components. A sub-pixel translation experiment is performed to avoid the affection of deformation and only consider the temporal performance. The original image was downloaded from the DIC challenge [28] as shown in **Fig. 3**. And a series of translated images were generated using Fourier transform. Also, random noise is added to analyze the robustness against noise. With a step of 0.05 pixels, 20 deformed images are simulated using Fourier transform with displacement following the formula:

$$u(t) = \frac{t}{20}\ pixels \tag{33}$$

Then, Gaussian noise of zero mean and different variances was added to the generated image series. Here, 5 noise levels range from 1% to 5% of the maximum grayscale (255 in 8-bit gray image) were added. After the image generation, the traditional spatial DIC method and the ST-DIC method described in 2.4 were used for comparison, where the former uses the first-order spatial shape function while the latter uses the first-order spatial and temporal shape function. In both methods, the spatial subset size is 31×31 pixels with 10 pixels step in both directions, and the calculation points are the same. Similar to the subset size, in ST-DIC, 5 frames in the temporal dimension are used. Both methods use the ZNSSD criterion and IC-GN optimization algorithm. In each frame, we use the mean L1-norm to evaluate the measurement error between calculated displacements and the given one:

$$error = \frac{1}{N} \| u_m - u \|_1 \tag{34}$$

where $N$ is the number of calculation points, $u_m$ is the measured displacement vector of all calculation points and $u$ is the given one.

The sub-pixel translation results are shown in **Fig. 4**. The left two subplots show the results of ST-DIC and the results of spatial DIC is shown in the right. In the zero-noise case, errors show sinusoidal trend w.r.t. the sub-pixel displacement in both methods. The errors and standard deviation (SD) are increasing along with the noise level as expected in both methods. In each level of noise, errors and SD using ST-DIC are lower than that of spatial DIC. And it can be seen intuitively that the result of processing 5% noise image using ST-DIC has the same level of error and SD as the result of processing 3% noise image using spatial DIC, both in errors and SD. It is convincing that the ST-DIC can further improve the robustness against image noise.

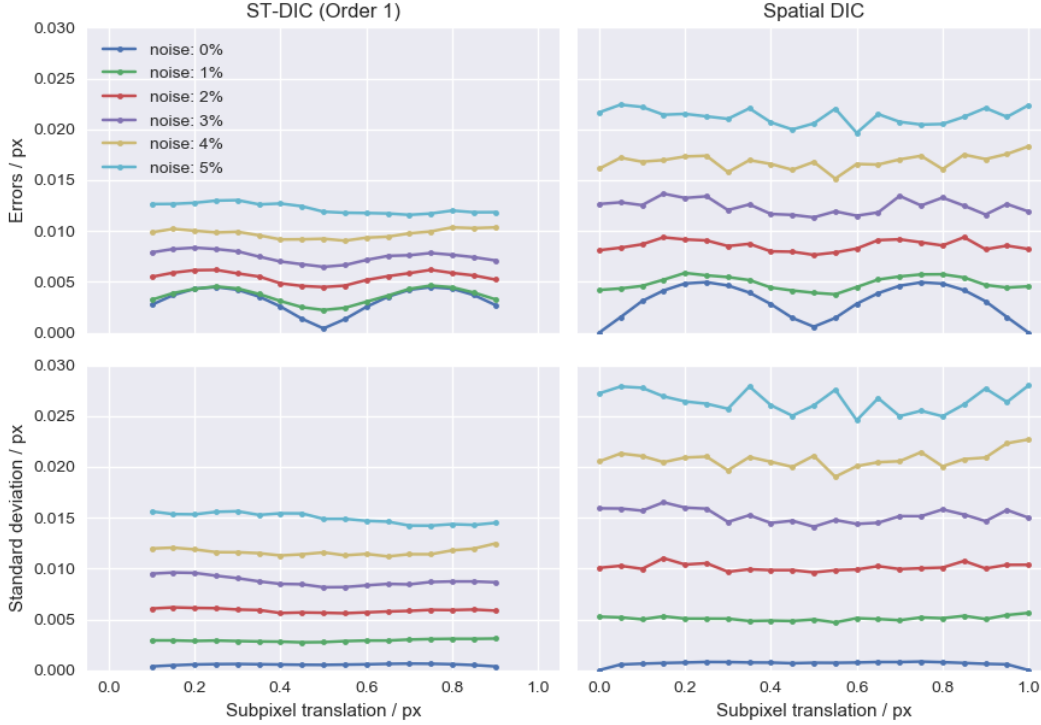

**Fig. 4** The sub-pixel translation errors and standard deviation using temporal and spatial DIC

In the sub-pixel translation experiment, the displacement and time are in a linear relationship. Hence, the first-order temporal shape function, which implies the linear velocity, can perfectly fit this situation.

### 3.1.2  Simulated vibration analysis

For real-world experiments, ST-DIC is used in image sequence analysis where the time step between adjacent image is a known constant. A typical example is to measure the vibration using a high-speed camera. In vibration testing, the displacements are rarely linear. And first-order temporal shape function will introduce the under-matched error as in spatial analysis [5]. To study the performance of ST-DIC in different velocity characters, we simulated two vibration displacements as $u$ and $v$ respectively, and applied them to the reference image to generate a series of deformation image, where:

$$u(t) = 10e^{-2t} \sin 10t$$
$$v(t) = 10e^{-3t} \sin 5t \qquad (35)$$

The imposed displacements are shown in **Fig. 5**. The time step is 0.01s and there are 200 frames in the 2s testing time span. It is clear that the distribution of sample points near the extreme point is not linear, and the local linearity goes better as time increases. The displacement mode combines the sharp and gentle changes. Similarly, noise of 5 levels varying from 1% to 5% was added to discuss the robustness against the noise. Here, we used three DIC algorithms to measure the given displacement: the simple first-order spatial DIC, the first-order ST-DIC described in 2.4 and the ST-DIC with second-order temporal shape function (parameters w.r.t. x: $[u, u_x, u_y, u_t, u_{tt}]$). Other factors remained unchanged.

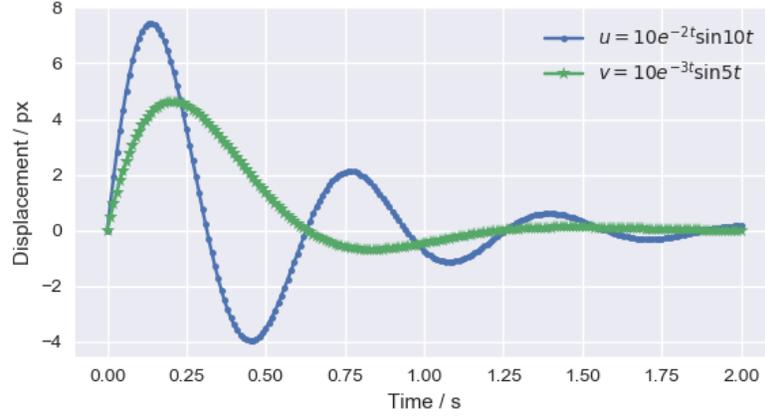

**Fig. 5** The imposed displacements u and v.

For intuitiveness, results of $u$ and $v$ are shown in **Fig. 6** and **Fig. 7**. For the sharply changed displacement $u$, the constant velocity assumption may not be satisfied near the extreme points. As shown in **Fig. 6**, in the first-order ST-DIC, the under-matched error dominated the bias error near the extreme points, performing much worse than the spatial method. Whereas, the second-order ST-DIC can fit the local non-linearity better and showed the best performance in the initial sharp changed situation ($t < 1$s). The method successfully overcame the under matched errors and suppressed the noise interference. When the displacement change became gentle ($t > 1$s), the first order method showed its superiority over the other two methods, the errors were less than the second-order method and showed less SD in the time axis.

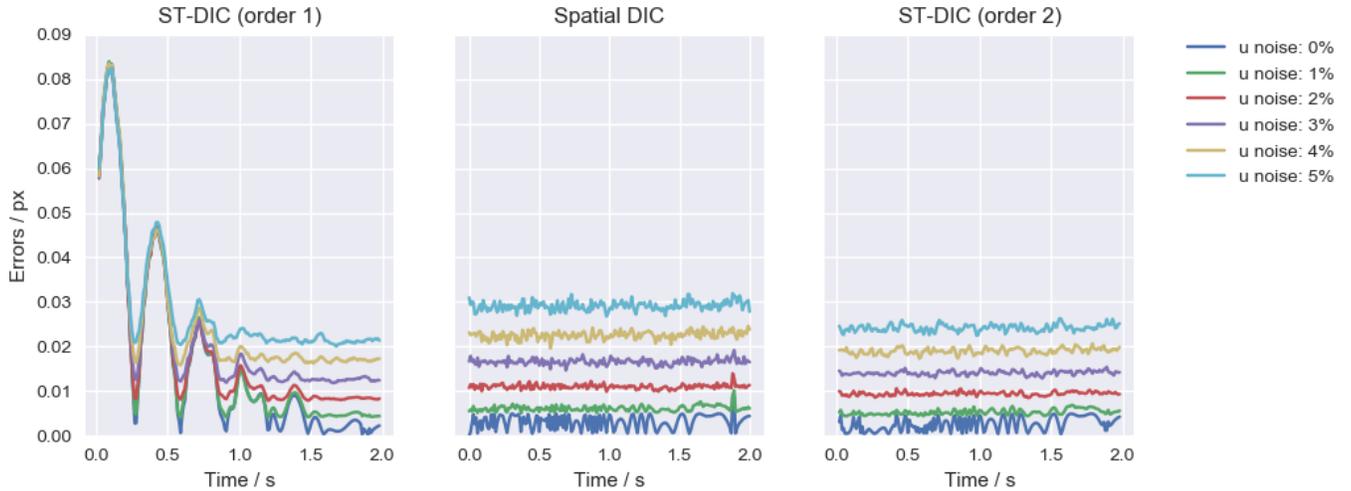

**Fig. 6** The measured u displacements using three DIC methods.

For the $v$ displacement, the results supported the same conclusion. The first-order ST-DIC preformed worse in the sharply changed area but showed the best in the gentle changed area. The second-order ST-DIC showed equally stability in both cases and have better robustness against noise than the classic spatial DIC under every noise level. In the flat case, when $t > 1s$, the ratio of mean errors along the time of other methods to the linear method are listed in **Table 1**. The mean errors of the second-order method are 10% to 15% higher than the linear method, while the errors of the traditional spatial method are 30% to 40% higher. It can be concluded that the second-order ST-DIC is suitable for both linear and nonlinear displacements due to its great performance on complex displacements and anti-noisiness.

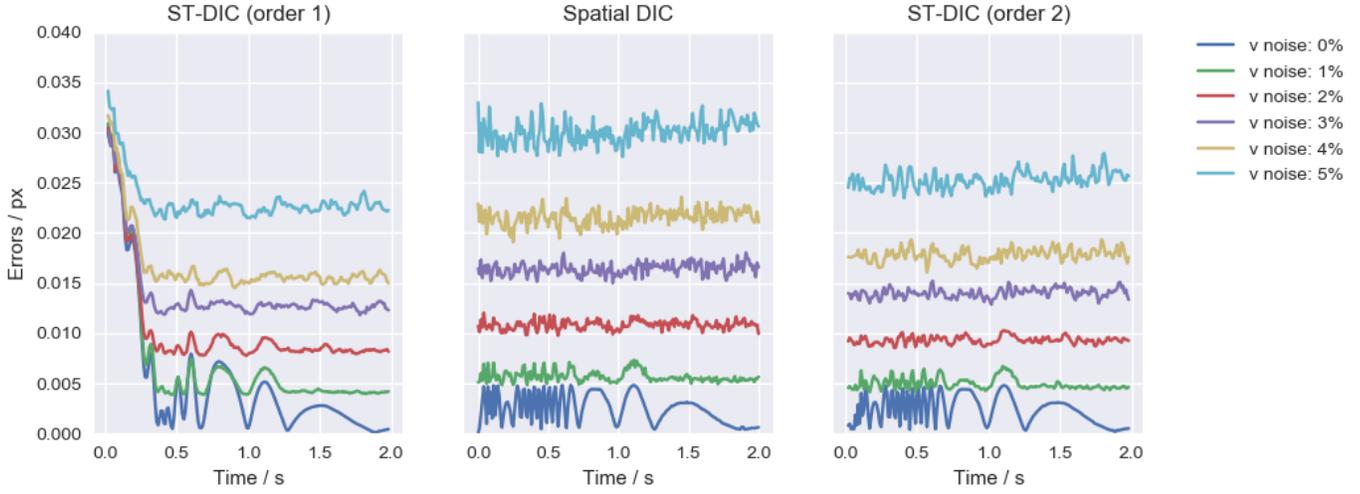

**Fig. 7** The measured v displacements using three DIC methods.

**Table 1** The mean error ratio (# / ST-DIC (order 1))

| noise / % | Spatial DIC | ST-DIC (order 2) |
| --- | --- | --- |
| 0 | 1.033 | 1.031 |
| 1 | 1.252 | 1.092 |
| 2 | 1.295 | 1.112 |
| 3 | 1.308 | 1.116 |
| 4 | 1.403 | 1.157 |
| 5 | 1.348 | 1.132 |

### 3.2  Application of mitigating air disturbance due to heat haze

As described in our previous work [19], the air disturbance due heat haze brings adverse effects in thermomechanical testing. The unsteady flow field between the camera and the specimen will bring huge noise to the deformation measurement. Because of this, the thermal expansion displacement field of isotropic material will be no longer concentric under the disturbance. Note that the air disturbance and the true deformation are coupled and cannot be easily separated. But in this experiment, the temperature of the specimen is applied by control but the air disturbance is random, the temporal-spatial method described in 2.6 can be used to apply the temporal constraint and mitigate the disturbance to some extent.

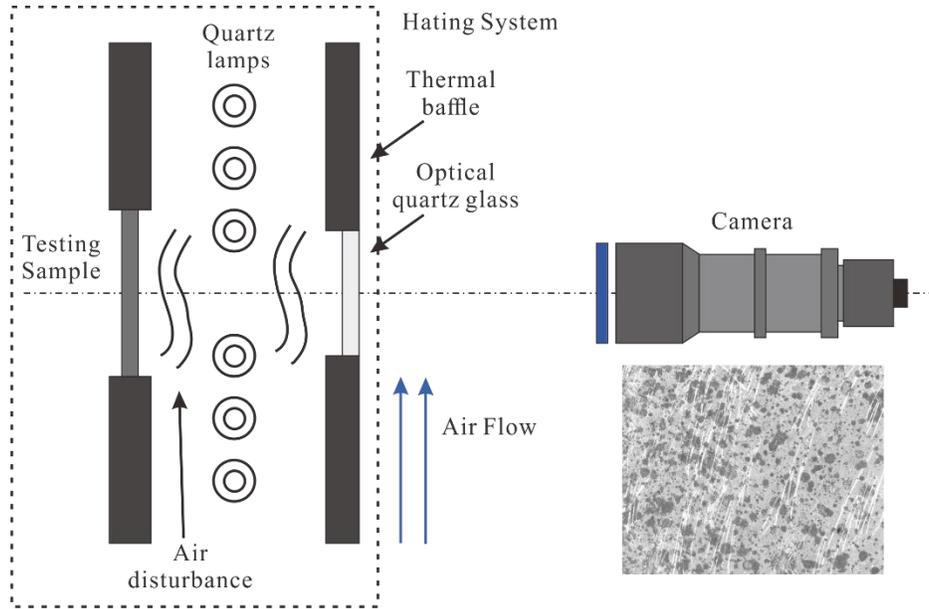

**Fig. 8** Schematic diagram of the experiment set-ups

Here, we just apply the proposed method to the recorded images in the experiments described in the previous work[19]. The experiment measures the coefficient of thermal expansion (CTE) using 2D-DIC. As shown in **Fig. 8**, quartz lamps are used to heat the front surface (observed surface) of the testing sample. Because the heating system is isolated from the camera, the camera can only observe the testing sample through the observation window made of optical quartz glass and the air flow can only reduce the air disturbance outside the heating system. The air disturbance within the heating system still exists. More details about the experiment can be found in [19]. Also, a sample of the captured images is shown in **Fig. 8** and a video intuitively showing the air disturbance is attached as supplementary material. These images are analyzed using two methods (first order spatial-DIC and first-order ST-DIC) based on the proposed framework. In both methods, the spatial subset sizes are $51 \times 51$ pixels and the grid steps are 15 pixels. In the spatial-temporal method, the temporal subset size is set to 5 to apply enough temporal constraints.

**Fig. 9** shows the displacement fields at different temperatures using the mentioned two methods. Note that the rigid body displacement has been removed. In this experiment measuring the CTE, the contours should be regular concentric circles in the ideal situation as the testing sample is expansion freely under the thermal load. As shown in the second row of **Fig. 9**, the effect of heat haze is obvious. The contours obtained using traditional spatial DIC is more deformed and farther away from concentric circles while that of ST-DIC shows much better displacement fields. Comparing with the contours obtained using ST-DIC, the traditional one shows a compression disturbance. These anisotropy results are due to the directionality of the quartz lamp. In the three typical temperatures, it is apparent that the use of temporal constraints can mitigate the disturbance due to heat haze.

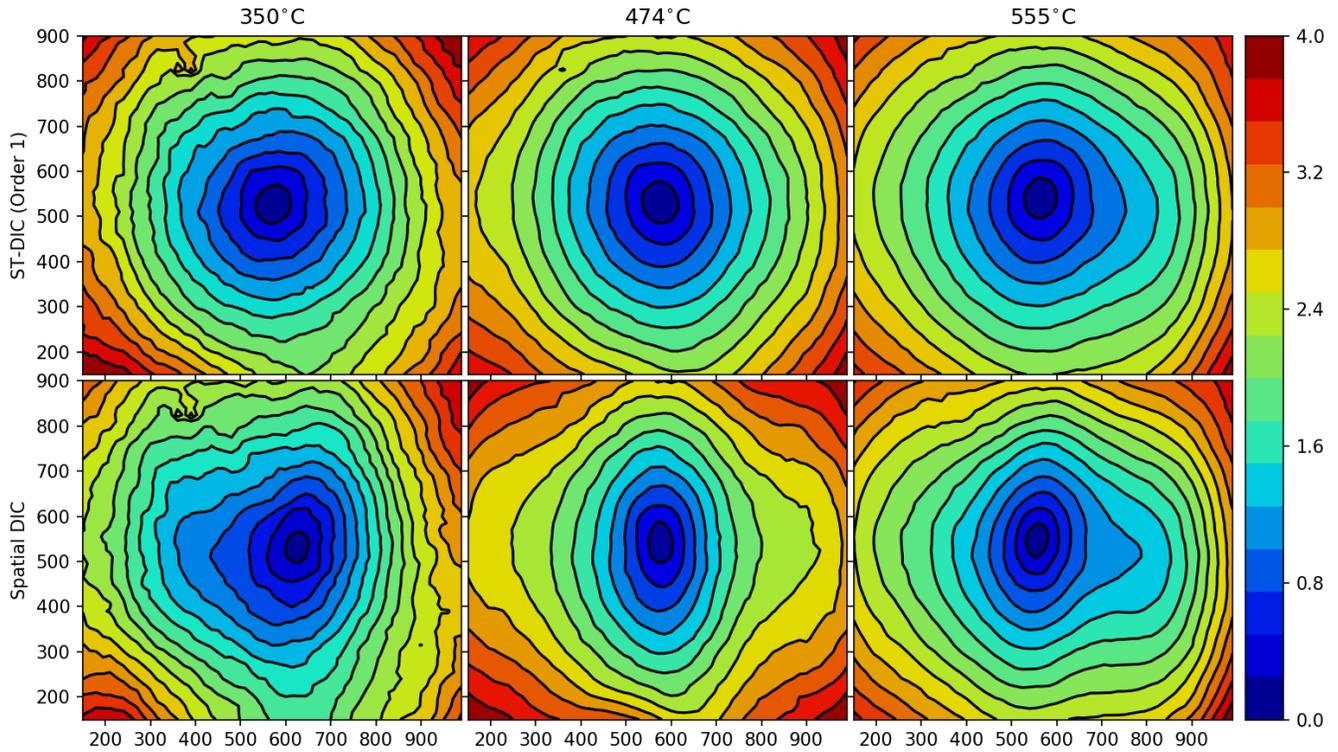

**Fig. 9** The displacement fields at different temperatures using temporal-spatial DIC (first row) and traditional spatial DIC (second row).

In the experiments, the strain along two directions should be equal and homogeneous within the ROI. **Fig. 10**

gives the mean strain components of $u_x$ and $v_y$ using two methods along with temperature. The mean strain obtained using traditional spatial-DIC is shown as dot while that of ST-DIC is shown in line. It is clear that the line goes through the dots in both x and y directions, indicating the results of ST-DIC has less fluctuation and higher linearity. Through linear fit, the CTE and the $R^2$ along x and y can be obtained and shown in **Table 2**. The difference of CTEs in two directions using ST-DIC is 0.06 $\mu\varepsilon/°C$ while that of spatial-DIC is 0.53 $\mu\varepsilon/°C$. The $R^2$ of the ST-DIC is higher than that of spatial-DIC in both directions. It can be concluded that the CTEs obtained using ST-DIC are of higher credibility. To further demonstrate the improvements of ST-DIC, the SD of the strain field along with temperature using two methods are shown in **Fig. 11**. The SD errors using ST-DIC are half of that of spatial-DIC in both directions, which means the strain fields are more homogeneous than that of spatial-DIC, indicating the robustness of ST-DIC against the air disturbance due to heat haze.

**Table 2** The CTE and $R^2$ of different methods

|  | Spatial DIC | ST-DIC |
| --- | --- | --- |
| $CTE_x$ ($\mu\varepsilon/°C$) | 19.66 | 19.04 |
| $R^2$ | 0.9884 | 0.9982 |
| $CTE_y$ ($\mu\varepsilon/°C$) | 19.13 | 18.98 |
| $R^2$ | 0.9975 | 0.9989 |

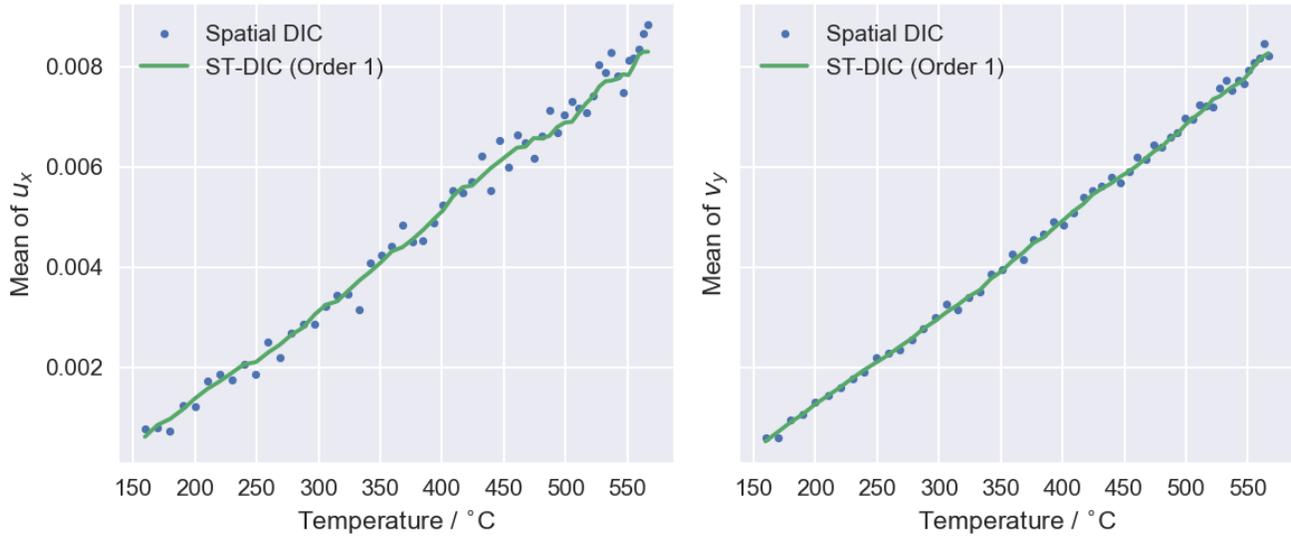

**Fig. 10** The mean strain components $u_x$ (left) and $v_y$ (right) using two methods at different temperatures.

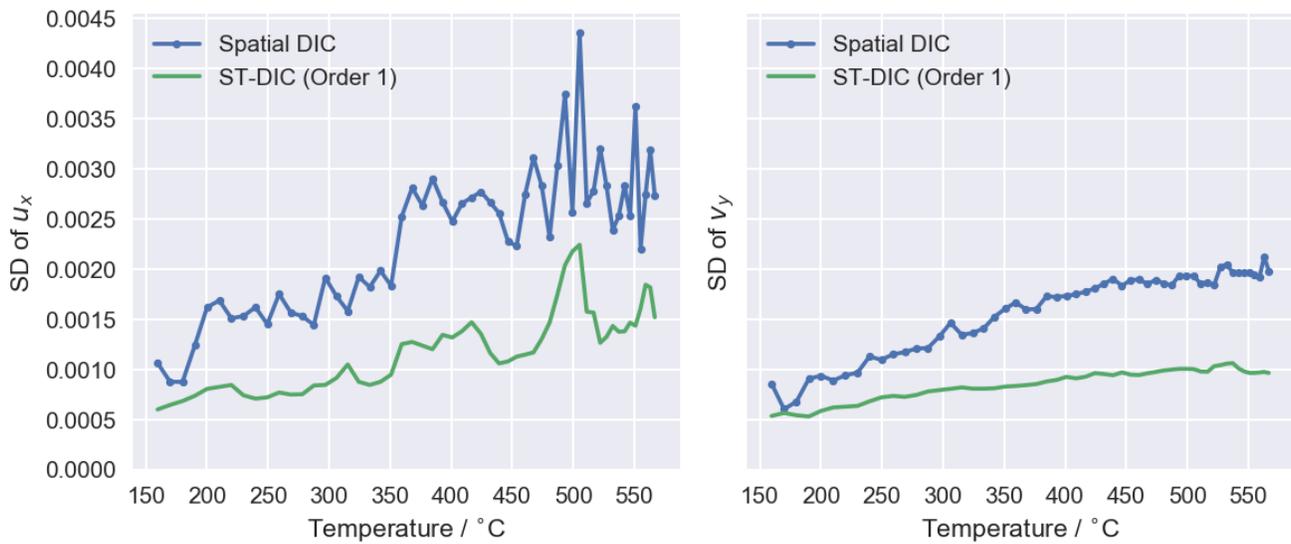

**Fig. 11** The mean strain components $u_x$ (left) and $v_y$ (right) using two methods at different temperatures.

## 4 Conclusions

In this paper, a comprehensive and systematic framework that can easily extend and implement the spatial-temporal DIC (ST-DIC) algorithm is proposed. The three main factors (optimization algorithms, correlation criterion and shape functions) in DIC problem are decoupled and discussed in detail. The framework unifies the time and space dimension and treats them as the same status. Hence, it can represent the traditional spatial DIC and ST-DIC in a uniform form. Researchers can freely choose the factors or add more parameters to form a new DIC algorithm. Though ZNSSD and IC-GN have become the first choice, the shape function can be modified in different situations.

Further, the 1-order temporal shape function considering the strain rate based on the proposed framework was presented as a tutorial. The algorithm is deduced in the uniform form and can be easily adapted from the existing code. Simulated sub-pixel translation experiments proved the robustness against noise of the temporal methods. And the simulated image sequence with different velocity characters showed the different application scenarios of different temporal methods. The researcher should select different temporal shape functions or parameters according to the motion characteristics of the object to be measured. The real-world experiment measuring the CTE shows the robustness of the ST-DIC against the air disturbance due to heat haze. Though the adverse effect cannot be totally eliminated, the proposed method can mitigate the disturbance to some extent.

## Acknowledgments

This work is supported by the National Natural Science Foundation of China (NSFC) (11872009, 11632010), the National Key Research and Development Program of China (2018YFB0703500).

## Appendix A

### A.1 Linear LSQ

When $\mathbf{r}(\mathbf{p})$ is a linear combination of the parameter vector $\mathbf{p}$, we call this case *linear least squares*. That is, $\mathbf{r}(\mathbf{p})$ can be expressed as:

$$\mathbf{r}(\mathbf{p}) = \mathbf{A}\mathbf{p} + \mathbf{b} \tag{A.1}$$

We obtain the quadratic form of the objective function:

$$f(\mathbf{p}) = \frac{1}{2} \|\mathbf{Ap} + \mathbf{b}\|^2 = \frac{1}{2}\mathbf{p}^T\mathbf{A}^T\mathbf{Ap} + \mathbf{b}^T\mathbf{Ap} + \frac{1}{2}\mathbf{b}^T\mathbf{b} \tag{A.2}$$

Differentiating both sides of (A.2) gives the optimality condition:

$$\nabla f(\mathbf{p}^*) = \mathbf{A}^T\mathbf{Ap}^* + \mathbf{A}^T\mathbf{b} = 0 \tag{A.3}$$

Solving (A.3) gives the closed-form solution of parameters in linear least squares problem:

$$\mathbf{p}^* = -(\mathbf{A}^T\mathbf{A})^{-1}\mathbf{A}^T\mathbf{b} \tag{A.4}$$

**A.2 Nonlinear LSQ**

In general, however, there is not often the case that the residual vector can be expressed as linear form. When $\mathbf{r}(\mathbf{p})$ is a nonlinear function, we consider the first-order Taylor expansion of the object function about the current parameter vector $\mathbf{p}$:

$$f(\mathbf{p} + \Delta\mathbf{p}) = \frac{1}{2}\|\nabla\mathbf{r}(\mathbf{p})\Delta\mathbf{p} + \mathbf{r}(\mathbf{p})\|^2 \tag{A.5}$$

where $\mathbf{r}(\mathbf{p})$ can be substituted by introducing the Jacobi matrix:

$$\nabla\mathbf{r}(\mathbf{p}) = [\nabla r_1(\mathbf{p}), \nabla r_1(\mathbf{p}), ..., \nabla r_m(\mathbf{p})]^T = \mathbf{J}(\mathbf{p}) \tag{A.6}$$

where $\mathbf{J}(\mathbf{p})$ is an m-by-n matrix, and the element $J_{ij}$ the partial derivative of $r_i$ w.r.t. $x_j$. Then the nonlinear least squares problem can be iteratively solved. The iterative sub-problem can be expressed as:

$$\arg\min_{\Delta\mathbf{p}\in\mathbb{R}^n} f(\mathbf{p} + \Delta\mathbf{p}) = \frac{1}{2}\|\mathbf{J}(\mathbf{p})\Delta\mathbf{p} + \mathbf{r}(\mathbf{p})\|^2 \tag{A.7}$$

Using the closed-form solution (A.4) gives the iterative updating parameter vector:

$$\Delta\mathbf{p}^* = -(\mathbf{J}(\mathbf{p})^T\mathbf{J}(\mathbf{p}))^{-1}\mathbf{J}(\mathbf{p})^T\mathbf{r}(\mathbf{p}) \tag{A.8}$$

then the iterative form can be formed as:

$$\mathbf{p}^{(k+1)} = \mathbf{p}^{(k)} - \mathbf{J}(\mathbf{p}^{(k)})^\dagger \mathbf{r}(\mathbf{p}^{(k)}) \tag{A.9}$$

where $J^{\dagger}$ is the pseudo-inverse of matrix J.

Repeating the iteration until convergence gives the optimized parameter vector $\mathbf{p}^*$. This is the well-known Gauss-Newton method, a straightforward method for nonlinear LSQ problem.